\documentclass[runningheads]{llncs}
\usepackage[T1]{fontenc}
\usepackage{graphicx}
\usepackage{multicol}
\usepackage{subcaption}
\usepackage{float}
\usepackage{amsmath}
\usepackage{caption}
\usepackage{booktabs}
\usepackage{hyperref}
\usepackage{cleveref}

\graphicspath{{figures/}} 

\usepackage{geometry}

\begin{document}
\title{Flood Data Analysis on SpaceNet 8 Using Apache Sedona}
\titlerunning{}
%

\author{Yanbing Bai\inst{1} \and Zihao Yang \inst{1} \and Jinze Yu\inst{2} \and Rui-Yang Ju\inst{3} \and Bin Yang\inst{4}  \and Erick Mas\inst{5}\and Shunichi Koshimura\inst{5}\thanks{Corresponding author: Shunichi Koshimura}}

\authorrunning{Yanbing Bai et al.} 
%
\tocauthor{Yanbing Bai}
%
\institute{Center for Applied Statistics, School of Statistics, Renmin University of China, Beijing 100872, China \and Connected Robotics Inc., Tokyo 184-0002 Japan \and Graduate Institute of Networking and Multimedia, National Taiwan University, Taipei City, 106335 Taiwan \and China Unicom Research Institute, Beijing 100048 China \and International Research Institute of Disaster Science, Tohoku University, Sendai 980-8572 Janpan\\ \email{koshimura@irides.tohoku.ac.jp}}

\maketitle              
\parindent=.25cm
\baselineskip=3.8truemm
\columnsep=.5truecm

\begin{abstract}
With the escalating frequency of floods posing persistent threats to human life and property, satellite remote sensing has emerged as an indispensable tool for monitoring flood hazards. SpaceNet8 offers a unique opportunity to leverage cutting-edge artificial intelligence technologies to assess these hazards. A significant contribution of this research is its application of Apache Sedona, an advanced platform specifically designed for the efficient and distributed processing of large-scale geospatial data. This platform aims to enhance the efficiency of error analysis, a critical aspect of improving flood damage detection accuracy. Based on Apache Sedona, we introduce a novel approach that addresses the challenges associated with inaccuracies in flood damage detection. This approach involves the retrieval of cases from historical flood events, the adaptation of these cases to current scenarios, and the revision of the model based on clustering algorithms to refine its performance. Through the replication of both the SpaceNet8 baseline and its top-performing models, we embark on a comprehensive error analysis. This analysis reveals several main sources of inaccuracies. To address these issues, we employ data visual interpretation and histogram equalization techniques, resulting in significant improvements in model metrics. After these enhancements, our indicators show a notable improvement, with precision up by 5\%, F1 score by 2.6\%, and IoU by 4.5\%. This work highlights the importance of advanced geospatial data processing tools, such as Apache Sedona. By improving the accuracy and efficiency of flood detection, this research contributes to safeguarding public safety and strengthening infrastructure resilience in flood-prone areas, making it a valuable addition to the field of remote sensing and disaster management.

\end{abstract}

\section{Introduction}
Floods stand as one of the most catastrophic natural disasters, inflicting significant loss of life, damage to infrastructure, and severe economic repercussions. The escalating frequency and intensity of flood events in recent years are largely attributed to climate change and urbanization. Rapid response during a flood event is crucial for saving lives and mitigating damage. Assessing the impact on buildings and infrastructure is essential for directing aid and identifying accessible routes. Satellite imagery analysis provides a fast initial mapping of flooded regions \cite{adriano2021learning}and aids in post-disaster monitoring. Mapping high flood risk areas, forecasting floods, and assessing post-flood effects are also key applications of remote sensing.

The development of flood detection models has been propelled by public benchmark datasets and challenges like SpaceNet. The SpaceNet Challenge series, initiated by a collaboration of companies, government agencies, and research institutions, has focused on building detection and road network extraction using high-resolution optical data. SpaceNet 8 \cite{hansch2022spacenet}, the latest challenge, aims to advance previous efforts by combining building and road annotations with flood detection, covering three Areas of Interest with approximately 32k buildings and 1,300km of roads, of which about 13\% and 15\% are flooded\cite{hansch2023spacenet}. This challenge seeks to establish a connection to previous challenges while introducing a realistic level of complexity, addressing automated mapping and flood detection in a unified framework.

Despite advancements in remote sensing and machine learning\cite{mueller2021monitoring}, flood detection models face significant challenges in efficiently analyzing and correcting errors, particularly in complex real-world scenarios. Misclassifications and inaccuracies can greatly impact model reliability and robustness. A key issue is how to effectively address errors when the model performs poorly. This involves examining instances of incorrect flood identification, understanding the underlying causes, and refining model performance. Common error patterns must be identified, and targeted improvements such as enhanced data preprocessing, additional feature incorporation, or model parameter fine-tuning need to be implemented. The ultimate goal is to improve the model's generalizability and accuracy across diverse geographic regions and environmental conditions, thereby enhancing its precision and robustness in real-world applications.

To address the challenge of efficiently analyzing and correcting errors in flood detection models, we introduce a novel approach that combines the strengths of Apache Sedona. Apache Sedona\cite{moussa2021scalable}, a distributed tool specifically designed for geospatial data processing, offers high efficiency in handling large-scale geospatial data. By leveraging Apache Sedona, we can accelerate the analysis of error cases, which is crucial for identifying the reasons behind poor model performance and enabling model improvement. Our methodology incorporates clustering algorithms to perform error case analysis, systematically identifying patterns and clusters of errors that can inform targeted corrections and model adjustments.

This research contributes to the field of remote sensing for flood detection by introducing a novel methodology that leverages the combined strengths of Apache Sedona. The main contributions lie on the following three points:

1. The paper introduces a novel approach that combines the usage of Apache Sedona, a distributed platform for geospatial data processing. This integration allows for efficient handling of large-scale geospatial data and leverages historical flood event cases to improve the accuracy and efficiency of flood damage detection.

2. The research shifts the focus from merely improving detection algorithms to analyzing the reasons behind inaccuracies in flood damage detection. This study systematically identifies and addresses the main sources of inaccuracies, such as flawed data annotations, low image contrast and models' capability, leading to targeted model optimization and enhanced precision in flood damage detection.

3. Employing clustering algorithms to systematically analyze error cases and refine the model is another contribution. This approach facilitates the identification of patterns and clusters of errors, enabling targeted corrections and model adjustments. By integrating this method, the research enhances decision-making and analysis quality in flood detection, contributing to more robust and effective flood detection models.

Our findings have the potential to advance satellite-based flood monitoring and contribute to sustainable development and disaster risk reduction efforts.
 
\section{Related Works}
\subsection{Spatial Computing System}
\subsubsection{Apache Sedona and Sedona with Apache Spark}

Apache Sedona is an open-source distributed computing library crafted for large-scale spatial data processing, extending Apache Spark for batch processing and Apache Flink for stream processing, with raster as the foundational image format. It provides an array of distributed spatial datasets and spatial SQL capabilities, enabling efficient distributed spatial data processing and analysis. The library's standout features include its speed, as benchmark tests and related academic papers suggest Sedona processes computationally intensive queries up to 10 times faster than other Spark-based geospatial data systems. Additionally, it boasts low memory consumption, with our benchmarks and third-party studies showing that Sedona's peak memory usage for large-scale in-memory query processing is 50\% less than that of other Spark-based geospatial data systems. Furthermore, its user-friendliness is highlighted by the provision of APIs for Scala, Java, spatial SQL, Python, and R, carefully integrated into the core system kernel, allowing for easy creation of spatial analysis and data mining applications that are cloud-environment compatible. Official use cases for Apache Sedona include Geographic Information Systems (GIS) for building maps and geographic data analysis, location intelligence and recommendation systems for analyzing locational data and providing recommendations, route planning, and location-based services, traffic management and planning for traffic flow data, route planning, and traffic forecasting, as well as environmental monitoring for processing spatial environmental data like meteorological and geological information.

``Sedona with Apache Spark'' represents Sedona's extension for batch processing within the Spark framework, delivering three distinctive functionalities: Firstly, it offers advanced spatial SQL capabilities supporting a wider range of spatial-related SQL query syntaxes, including Range Query, KNN Query, and Join Query based on spatial logic. Secondly, it provides various types of Spatial RDDs tailored to specific purposes and contexts, such as PointRDD for point data, PolygonRDD for polygon data, LinestringRDD for line data, CircleRDD for circular data, and RectangleRDD for rectangular data. Thirdly, Sedona's Python interface facilitates the creation and conversion between Sedona DataFrame and GeoDataFrame, which simplifies the visualization of geometric data.

\subsection{Deep learning model analysis}
Recently, the use of computer vision techniques to extract flood and building disaster information from satellite remote sensing imagery has become a research hotspot\cite{rudner2019multi3net,doshi2018satellite,doshi2018satellite,bonafilia2020sen1floods11,brunschwiler2023sustainable,li2024glh,fraccaro2022deploying,xu2024spatial}. The application of this technology can not only help us respond more quickly to natural disasters but also play an important role in post-disaster reconstruction.

For example, Fraccaro et al.\cite{fraccaro2022deploying} developed an artificial intelligence application that utilizes Sentinel 1 data to detect floods. This method identifies areas affected by floods by analyzing satellite data, which is crucial for disaster response and rescue efforts. Similarly, Xu et al.\cite{xu2019building} used convolutional neural networks (CNNs) to detect building damage in satellite imagery. This work identifies the affected conditions by analyzing the structural changes in buildings, which is valuable for post-disaster assessment and reconstruction planning. Furthermore, Mueller et al.\cite{mueller2021monitoring} employed machine learning techniques to monitor war-induced destruction. This research provides strong support for peace and reconstruction in conflict zones by analyzing the signs of destruction in satellite images.

With the release of the SpaceNet 8 dataset, researchers have been afforded a unique opportunity to study flood detection. Hänsch et al.\cite{hansch2022spacenet} used this dataset to detect flooded roads and buildings. This work provides an accurate assessment of the extent and severity of flood impact by analyzing pre- and post-flood satellite imagery. Although these studies have achieved significant technical results, they mainly focus on model accuracy and often neglect the analysis and reasoning of the results. Deep learning, as a commonly used technique, is considered a ``black box'', making it difficult to interpret its decision-making process. In this context, methods based on Apache Sedona offer a new perspective. By retrieving and analyzing specific cases, the methods can help us understand the logic behind deep learning model decisions or reason out new analysis methods to explain the model. Analyzing the incorrect judgments of deep learning models in specific situations can help better understand the limitations of the model and propose improvement strategies. This not only helps to improve the accuracy of the model but also enhances the explainability and transparency of the model, thereby playing a greater role in disaster response and sustainable development.

\section{Method}
\begin{figure}[t]
\centerline{\includegraphics[width=\columnwidth, keepaspectratio]{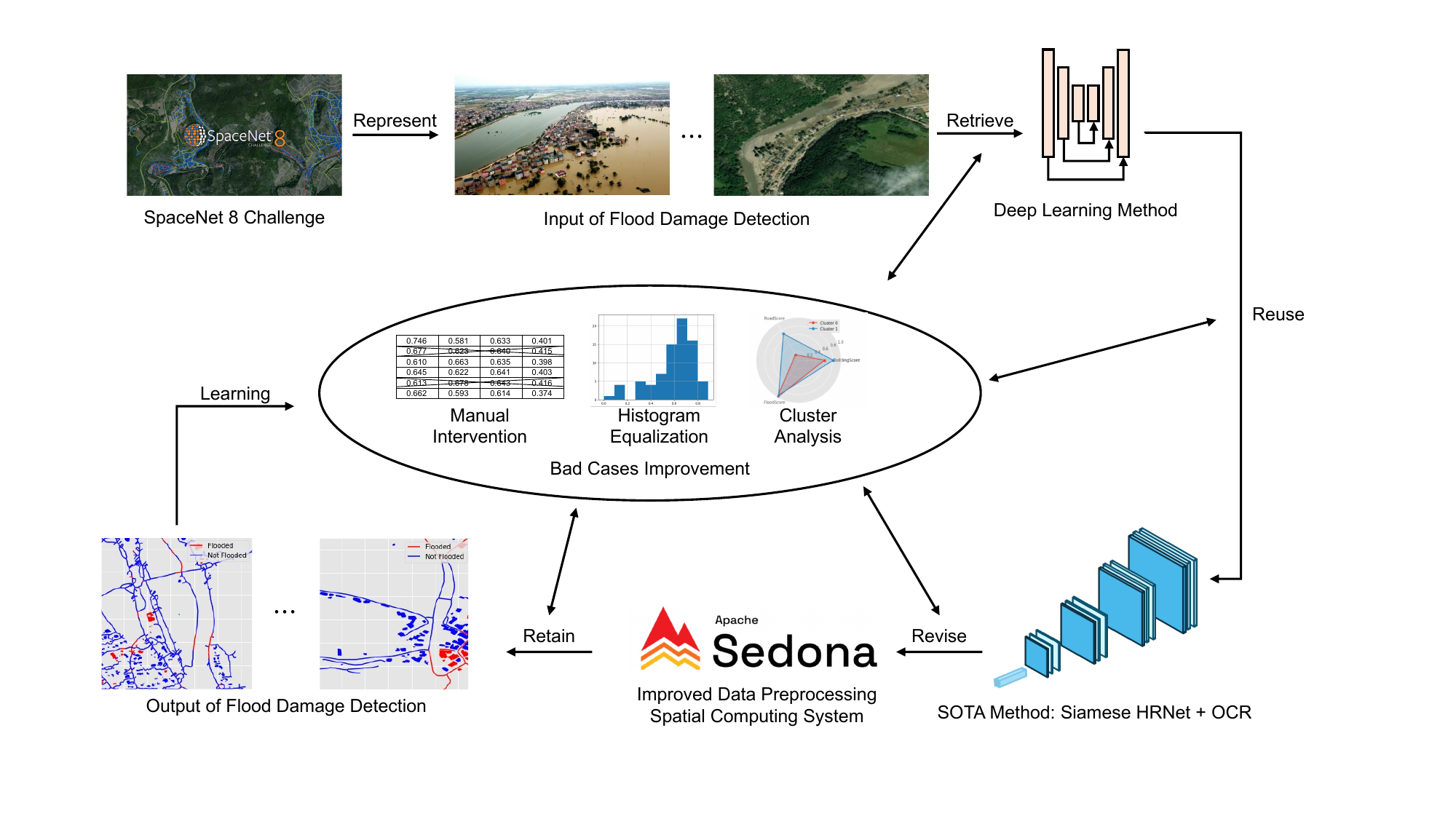}}
\caption{Overview of Our Work}
\label{fig:system}
\end{figure}

We propose a framework based on improved geospatial data analysis using Apache Sedona for flood detection, as shown in Figure \ref{fig:system}, which consists of which consists of Spatial Computing System, CBR and Deep Learning Methods \cite{koch2015siamese} parts.

By analyzing the influence of flood disasters on constructions and roads, flood disaster cases recorded in historical data to this situation can be extracted.\cite{wang2022predicting} The Apache Sedona big data platform for geospatial data can help us to quickly locate historical cases for this research case. This allows some inferences to be made from flood damage patterns and the more susceptible areas where damage occurs. Apache Sedona is an open-source distributed computing platform implemented for big-scale spatial data processing, on top of Apache Spark for batch processing and Apache Flink for stream processing, with raster as the underpinning of the image. Apache Sedona provides an accumulation of a variety of dispersed spatial datasets and distributed spatial-sql capabilities for efficient distributed spatial data processing and analysis.

Analyzing historical cases helps researchers understand typical damage to buildings and roads in flood disasters. This information can be adapted to current case analysis to predict specific damage, allowing for tailored models and strategies for each flood event and affected infrastructure. In our study, we preprocessed GeoJSON files by isolating road and building information, removing invalid records, and standardizing the format. We assigned speed limits to roads based on SpaceNet3 and SpaceNet5 methodologies. The cleaned GeoJSON files were exported as Shapefile files for GIS processing.

For deep learning masks construction, we buffered preprocessed Shapefile objects to create regions around them. This resulted in four mask types: \texttt{binary\_road}, \texttt{binary\_building}, \texttt{flood}, and \texttt{road\_speed}. The flood mask was divided into four channels: \texttt{non-flooded buildings}, \texttt{flooded buildings}, \texttt{non-flooded roads}, and \texttt{flooded roads}. The replication time for this step was approximately 2 hours on an NVIDIA Tesla P100-16GB.

After applying solutions from historical cases to the current problem, it may be necessary to revise the solution based on the specifics of the new case. For example, using clustering algorithms for misclassification analysis helps identify outliers, allowing researchers to adjust the model to more accurately reflect the impact of flood disasters on buildings and roads. U-Net, proposed by Olaf Ronneberger et al. in 2015, is a deep learning architecture\cite{adriano2023developing} widely used for semantic segmentation tasks. It has a U-shaped symmetric network structure with symmetric encoders and decoders connected by skip connections. The encoder performs down-sampling to extract features, while the decoder performs up-sampling to recover pixels. Skip connections concatenate features at different scales, combining shallow features with more spatial information and deep features with more semantic information, making U-Net effective for semantic segmentation tasks. Siamese U-Net, an extension of U-Net, is designed for processing paired image inputs and producing corresponding segmentation outcomes. It features two identical U-Nets that share weights, allowing the network to learn relationships between the two input images, such as changes or temporal variations in remote sensing imagery.

Upon the completion of the research, new case data (including problem descriptions, solutions, and results) can be stored in the case base.\cite{zhao2021spatial} This not only enriches the case base but also provides valuable resources for future research encountering similar problems. By maintaining a comprehensive and up-to-date case base, we can ensure that our models and strategies remain relevant and effective in addressing the evolving nature of flood disasters. High-Resolution (HR-Nwr) is a deep learning model for image recognition, proposed for the task of 2D human pose estimation. It significantly enhances network performance by maintaining high-resolution feature maps while concurrently extracting multi-level information. Traditional convolutional neural networks usually reduce the resolution of feature maps in the early stages of the network through downsampling operations (such as pooling or convolutions with large strides), resulting in a loss of detail information. HRNet maintains high-resolution representations throughout the process, starting with a high-resolution subnetwork as the first stage and gradually adding subnetworks from high to low resolution to form additional stages. It employs a Spatial Pyramid Pooling (SPP) module and connects multiple resolution subnetworks in parallel to capture multi-scale contextual information, performing well in semantic segmentation tasks that require the preservation of more detailed information.

HRNet-OCR incorporates self-attention architecture into HRNet, which enhances pixel representation by aggregating the similarity relationships between the target region and all object regions through weighted aggregation after obtaining coarse segmentation results from HRNet, thereby improving the model performance.

\section{Experiment}
\subsection{Dataset}
\subsubsection{SpaceNet8}
The SpaceNet8 dataset represents a significant advancement in remote sensing and machine learning, being the first to integrate building footprint detection, road network extraction, and flood identification \cite{ritwik2019xbd}. It includes high-resolution optical imagery from Maxar, captured before and after flood events in Germany, Louisiana, USA, and a blind test location. The dataset provides precise annotations for buildings, roads, and flood attributes in GeoJSON files, pre- and post-flood satellite images in TIFF files, and reference tables for road speeds \cite{etten2020city}.

SpaceNet8 features dehazed RGB imagery with a resolution of 0.3 to 0.8 meters, covering diverse areas such as Germany's hilly, riverside villages and Louisiana's flat, densely populated riverbanks. The dataset spans over 850 square kilometers, including 30,000 buildings and 1,300 kilometers of roads. It has been thoroughly annotated manually, with analysts marking every building and highway centerline, assessing post-flood conditions, and categorizing roads \cite{arndt2022spacenet}. This dataset is set to significantly advance flood detection, offering a valuable resource for developing and testing machine learning algorithms in remote sensing applications \cite{fraccaro2022deploying}.

\subsection{Data Preprocessing}
In our study, we preprocessed GeoJSON files by isolating road and building information, removing invalid records, and standardizing the format. We assigned speed limits to roads using SpaceNet3 and SpaceNet5 methodologies. The cleaned GeoJSON files were exported as Shapefile files for GIS processing, with the preprocessing time under 5 seconds on an NVIDIA Tesla P100-16GB.

We constructed deep learning masks by buffering preprocessed Shapefile objects, creating regions around the original geometric objects. This resulted in four mask types: \texttt{binary\_road}, \texttt{binary\_building}, \texttt{flood}, and \texttt{road\_speed}. The flood mask was divided into four channels: \texttt{non-flooded buildings}, \texttt{flooded buildings}, \texttt{non-flooded roads}, and \texttt{flooded roads}. The replication time for this step was approximately 2 hours on an NVIDIA Tesla P100-16GB.

Using Apache Sedona, we implemented spatial similarity queries to identify regions with similar topographical features or historical events, such as floods or earthquakes. This approach allowed us to apply insights from past flood events to current scenarios, providing a more informed basis for flood detection and response.

Lastly, we created training and validation sets by randomly splitting the data in a ratio of 0.85:0.15, with the replication time under 5 seconds on an NVIDIA Tesla P100-16GB. Based on Apache Sedona, we aim to develop a more robust and effective flood detection model that can adapt to real-world flood scenarios.

\subsection{Experimental Setup}
The Baseline model, based on data cleaning outcomes, comprises two independently trained convolutional neural networks (CNNs) \cite{xu2019building} with post-processing steps. The Foundational Features Network identifies buildings and roads using a ResNet34 U-Net model with binary cross-entropy loss for building segmentation and a combination of focal loss and soft-dice loss for road speed segmentation, plus a custom weighted loss function\cite{zhao2019region}. The Flood Attribution Network predicts flood status using a siamese U-Net model with cross-entropy loss. Post-processing transforms the network's output into vector data for building masks and road networks, with flood attributes determined by majority voting. The Baseline model trains efficiently, taking 4.5 minutes per epoch on an NVIDIA Tesla P100-16GB GPU.

\begin{table}[t]
\centering
\caption{SpaceNet8 Dataset: Areas of Interest Analysis}
\label{tab:spacenet8}
\setlength{\tabcolsep}{1.5mm}{
\begin{tabular}{cccccccc}
\noalign{\smallskip} \hline \noalign{\smallskip}
\textbf{Metric} & \begin{tabular}[c]{@{}c@{}}Area\\ (km\textsuperscript{2})\end{tabular} & \begin{tabular}[c]{@{}c@{}}No.\\ Bldgs\end{tabular} & \begin{tabular}[c]{@{}c@{}}Roads\\ (km)\end{tabular} & \begin{tabular}[c]{@{}c@{}}Bldg.\\ Inund.\end{tabular} & \begin{tabular}[c]{@{}c@{}}Inund. \\ Ratio \\ (\%)\end{tabular} & \begin{tabular}[c]{@{}c@{}}Road \\ Inund.\\ (km)\end{tabular} & \begin{tabular}[c]{@{}c@{}}Inund. \\ Ratio\\ (\%)\end{tabular} \\ \noalign{\smallskip} \hline \noalign{\smallskip} \hline \noalign{\smallskip}
\textbf{Germany} & 741.1 & 18,892 & 885 & 2,736 & 14 & 148.4 & 17 \\
\textbf{Louisiana} & 65.5 & 3,727 & 163.6 & 1,175 & 32 & 34.6 & 21 \\
\textbf{Blind Test Area} & 43.4 & 9,000 & 287.7 & 344 & 4 & 21.2 & 7 \\ \noalign{\smallskip} \hline
\end{tabular}}
\end{table}

The Top1 model improves upon the Baseline by refining the neural network architecture, specifically using a Siamese HRNet+OCR model with RMI loss and an auxiliary head (\text{aux\_head}) for enhanced prediction accuracy. The OCR module with an attention mechanism updates pixel representations, improving contextual information capture and overall performance. The Top1 model's training speed is 7.5 minutes per epoch on 2 NVIDIA Tesla P100-16GB GPUs, with 100 epochs and a batch size of 4.

In the realm of deep learning, we use loss functions to guide the optimization process and ensure the effectiveness of our models. Each loss function serves a distinct purpose and is tailored to specific types of problems, thereby playing a crucial role in the training and performance of our neural networks.

\subsection{Experimental Results}
For the SpaceNet8 challenge, the official organizers provided two key fundamental metrics for assessing the quality of solutions. The first metric is the Intersection over Union (IoU), which quantifies the effectiveness of building flood detection. The building status includes no building (based on a fixed area as the standard), flooded building, and non-flooded building. The metric is described using a 3x3 matrix with rows representing reference data situations, columns representing predicted situations, and cells containing the average IoU, thereby describing the related states of building objects. The overall quality score is calculated based on this matrix, and the effects of different solutions are compared. The second metric is the Average Path Length Similarity (APLS), which quantifies the predicted effectiveness of road network travel time or distance. It is an indicator based on graph theory that describes the similarity of network structures. The APLS calculates the similarity of graphs by computing the average path length and its permutation similarity for two states: roads that are flooded and roads that are not.

\begin{figure}[t]
\centering
\renewcommand\thesubfigure{\alph{subfigure}} 
\begin{subfigure}[a]{0.2\linewidth}
\includegraphics[width=\linewidth]{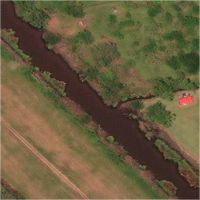}
\captionsetup{justification=centering}
\caption{Target Omission\\(True label)}
\label{fig:sub1a}
\end{subfigure}
\quad
\setcounter{subfigure}{1}
\begin{subfigure}[]{0.2\linewidth}
\includegraphics[width=\linewidth]{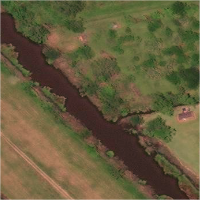}
\captionsetup{justification=centering}
\caption{Target Omission\\(Pred label)}
\label{fig:sub1b}
\end{subfigure}
\quad
\setcounter{subfigure}{2}
\begin{subfigure}[]{0.2\linewidth}
\includegraphics[width=\linewidth]{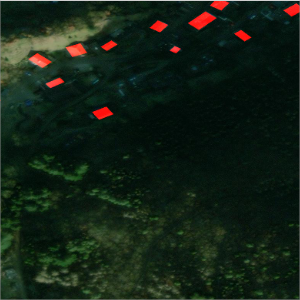}
\captionsetup{justification=centering}
\caption{Spatial Confusion\\(True label)}
\label{fig:sub2a}
\end{subfigure}
\quad
\setcounter{subfigure}{3}
\begin{subfigure}[]{0.2\linewidth}
\includegraphics[width=\linewidth]{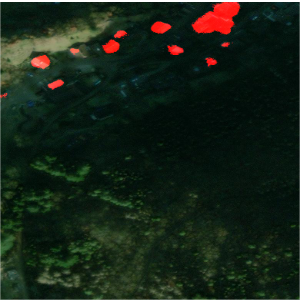}
\captionsetup{justification=centering}
\caption{Spatial Confusion\\(Pred label)}
\label{fig:sub2b}
\end{subfigure}

\begin{subfigure}[b]{0.2\linewidth}
\includegraphics[width=\linewidth]{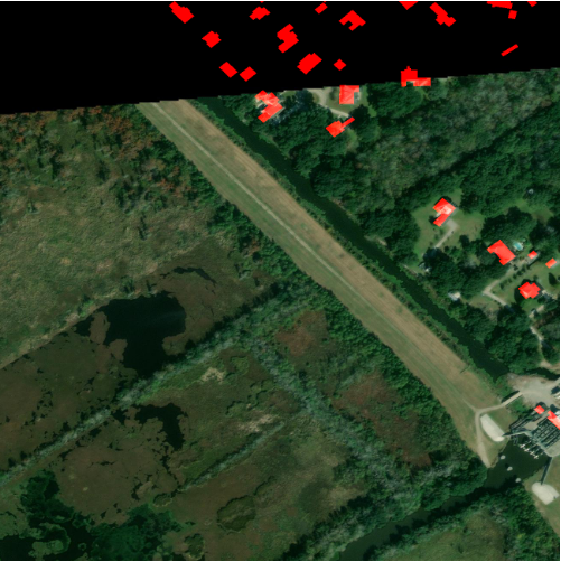}
\captionsetup{justification=centering}
\caption{Missing Information\\(True label)}
\label{fig:sub3a}
\end{subfigure}
\quad
\setcounter{subfigure}{5}
\begin{subfigure}[b]{0.2\linewidth}
\includegraphics[width=\linewidth]{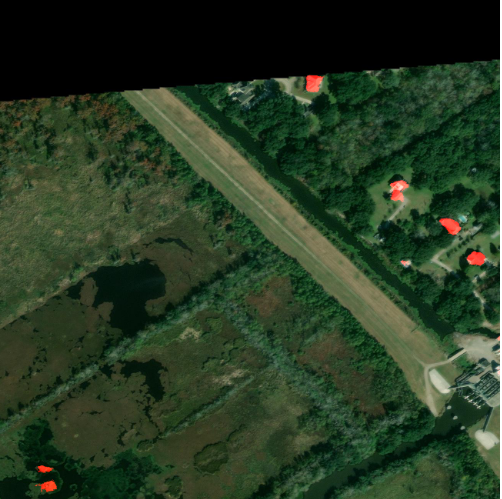}
\captionsetup{justification=centering}
\caption{Missing Information\\(Pred label)}
\label{fig:sub3b}
\end{subfigure}
\quad
\setcounter{subfigure}{6}
\begin{subfigure}[b]{0.2\linewidth}
\includegraphics[width=\linewidth]{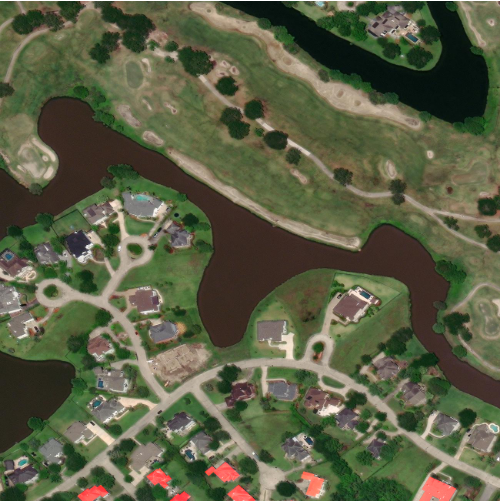}
\captionsetup{justification=centering}
\caption{Inherent Inaccuracy\\(True label)}
\label{fig:sub4a}
\end{subfigure}
\quad
\setcounter{subfigure}{7}
\begin{subfigure}[b]{0.2\linewidth}
\includegraphics[width=\linewidth]{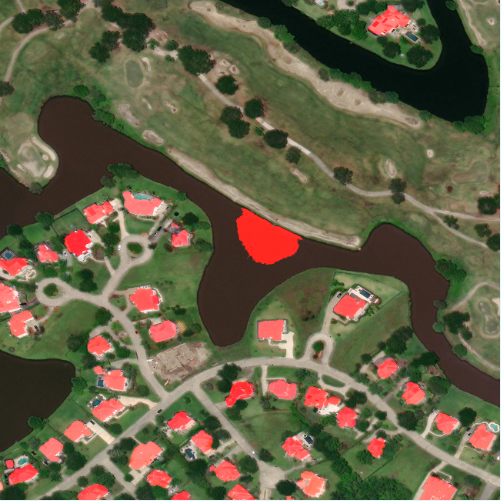}
\captionsetup{justification=centering}
\caption{Inherent Inaccuracy\\(Pred label)}
\label{fig:sub4b}
\end{subfigure}

\caption{The model identified four primary error types: (a)(b) Target omission due to small object size; (c)(d) Spatial confusion, where targets blend with surroundings; (e)(f) Missing information from incomplete satellite imagery labels; (g)(h) Incorrect labels from manual annotation errors. The illustration shows target objects (red areas), actual labels (``True label''), and model predictions (``Pred label'').}
\label{fig:errors}
\end{figure}

The most significant innovation of this study lies in conducting error analysis based on existing approaches. Firstly, the study identifies errors using the Intersection over Union (IoU) as the primary model evaluation criterion (a pixel-level assessment metric suitable for raster-based data formats). Secondly, the Apache Sedona platform is utilized to visualize the errors and summarize the main types and inferred causes of errors. Lastly, the study improves the approach based on the error analysis. 

The results of the error analysis reveal that, on one hand, incorrect label information in the dataset affects the objective evaluation of model effectiveness; that is, when information is missing or labels are incorrect, predictions that are originally correct are judged as errors, leading to an underestimation of model performance. On the other hand, there is still room for improvement in the model's recognition capability; it is possible to further reduce cases of target omission and spatial confusion errors by refining the approach. This leads to Experiment 2 of this study: improving the flood detection scheme in satellite remote sensing images using building recognition as an example.

The results of the error analysis indicate that, on one hand, incorrect label information in the dataset affects the objective evaluation of model performance. Specifically, when information is missing or labels are incorrect, predictions that are originally correct will be judged as errors, leading to an underestimation of model effectiveness. On the other hand, there is still room for improvement in the model's recognition capability. Efforts can be made to further reduce instances of target omission and spatial confusion errors through improved approaches. This sets the stage for Experiment 2 of this study: using building recognition as an example to enhance the flood detection scheme in satellite remote sensing images.

\subsection{Experiment 1: Cluster Analysis}
To enhance our understanding and application of error analysis in the context of flood detection using deep learning models\cite{corbat2020fusion}, we conducted three experiments. These experiments are designed to explore different aspects of model performance and to identify areas for improvement. Our first experiment delves into cluster analysis, where we scrutinize the scores obtained from both the Baseline and State-of-the-Art(SOTA) models. The objective is to discern feature differences between various clusters, thereby shedding light on the models' abilities to predict different image features. Furthermore, this analysis aids in identifying image features that may lead to suboptimal prediction outcomes, offering valuable insights for model refinement. 

\begin{figure}[t]
\centering
\includegraphics[width=\textwidth]{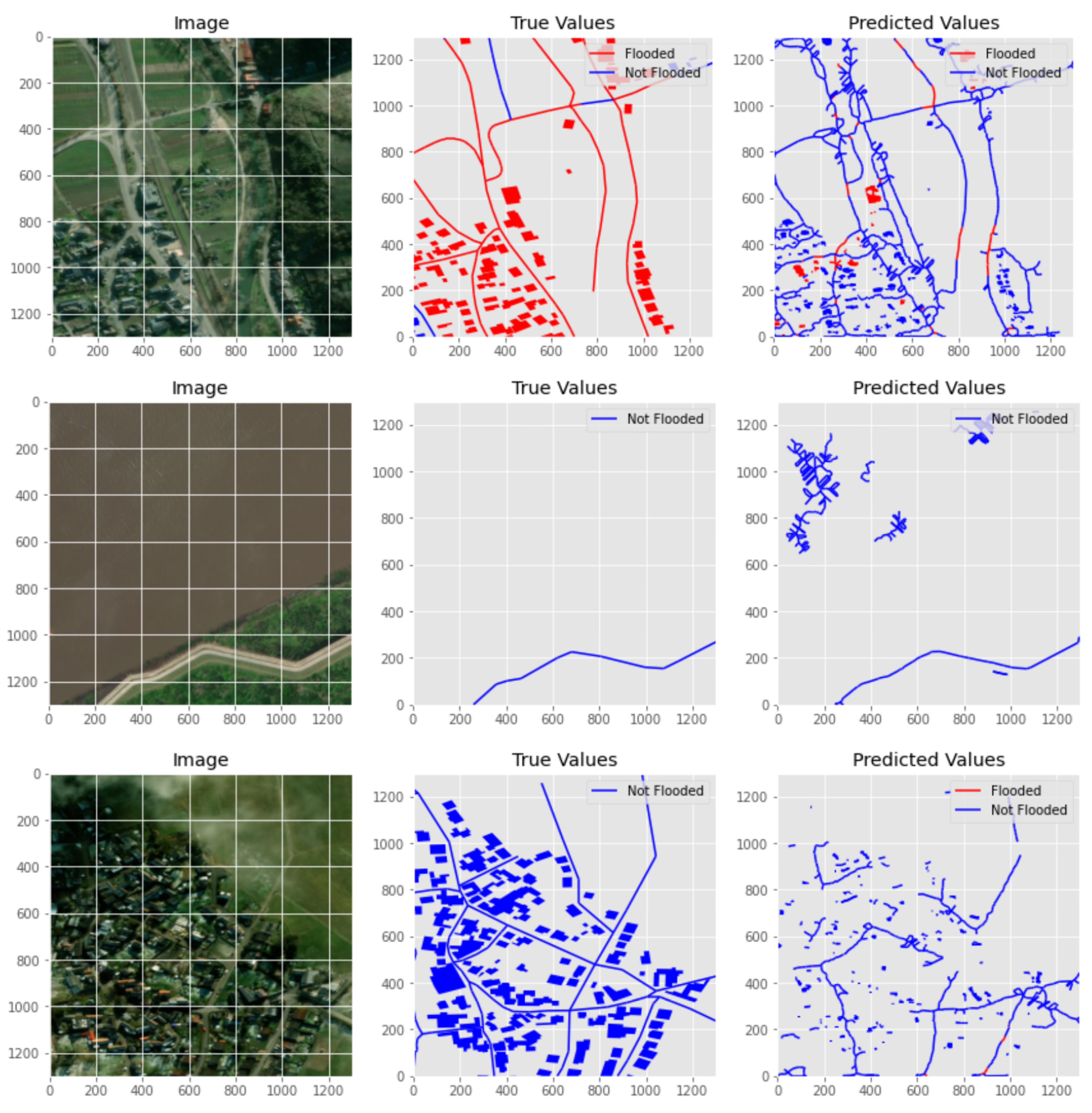}
\caption{Comparison between satellite images, annotated images, and Baseline prediction images. From top to bottom are examples of \textbf{Cluster 0}, \textbf{Cluster 1}, and \textbf{Cluster 2}.}
\label{compar_cluster_baseline}
\end{figure}

We conduct a cluster analysis of the scores obtained from the Baseline and SOTA model, aiming to observe feature differences between different clusters, thereby revealing the models' predictive capabilities for different image feature. Besides, we can also identify image features that may lead to poorer prediction outcomes, then provide valuable guidance for model improvement .And the comparison results are shown in Figure \ref{compar_cluster_baseline}.

By observing this figure, we can summarize the characteristics of the three clusters as follows. \textbf{Cluster 0} demonstrates relatively balanced scores in roads, buildings, and floods. However, its flood prediction score is notably lower, deviating from the ideal value of 1. According to Figure \ref{compar_cluster_baseline}, samples in this cluster exhibit extremely poor performance in flood prediction. This outcome may suggest significant flaws or limitations in the model's ability to recognize flood features. \textbf{Cluster 1} scores close to 1 in floods and buildings, while relatively lower in roads. Upon observation of Figure \ref{compar_cluster_baseline}, this cluster often predicts roads that do not actually exist, which may not solely be attributed to limitations of the model itself, but could also be related to incomplete annotations in the original annotated data. This finding underscores strict management and control of data quality can improve model predictive performance and generalization ability. \textbf{Cluster 2} scores close to 1 in flood prediction, while its scores for roads and buildings are relatively lower. As shown in Figure \ref{compar_cluster_baseline}, the predictions indicates the model's capability in identifying flood features to some extent. However, further analysis reveals that the model may fail to accurately capture roads and buildings in the images, which result in its difficulties or biases in the model's recognition of roads and buildings features.

\begin{figure}[t]
\centering
\renewcommand\thesubfigure{\alph{subfigure}} 
\begin{subfigure}[a]{0.3\linewidth}
\includegraphics[width=\linewidth]{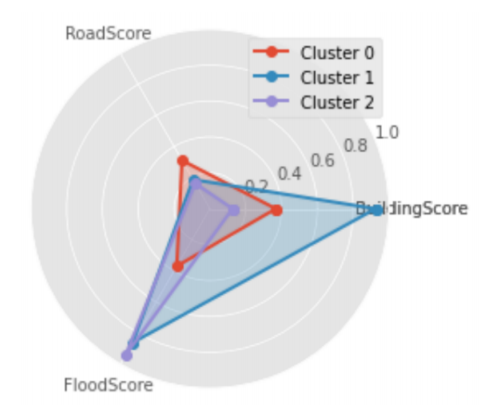}
\captionsetup{justification=centering}
\caption{Baseline Model}
\end{subfigure}
\quad
\setcounter{subfigure}{1}
\begin{subfigure}[]{0.3\linewidth}
\includegraphics[width=\linewidth]{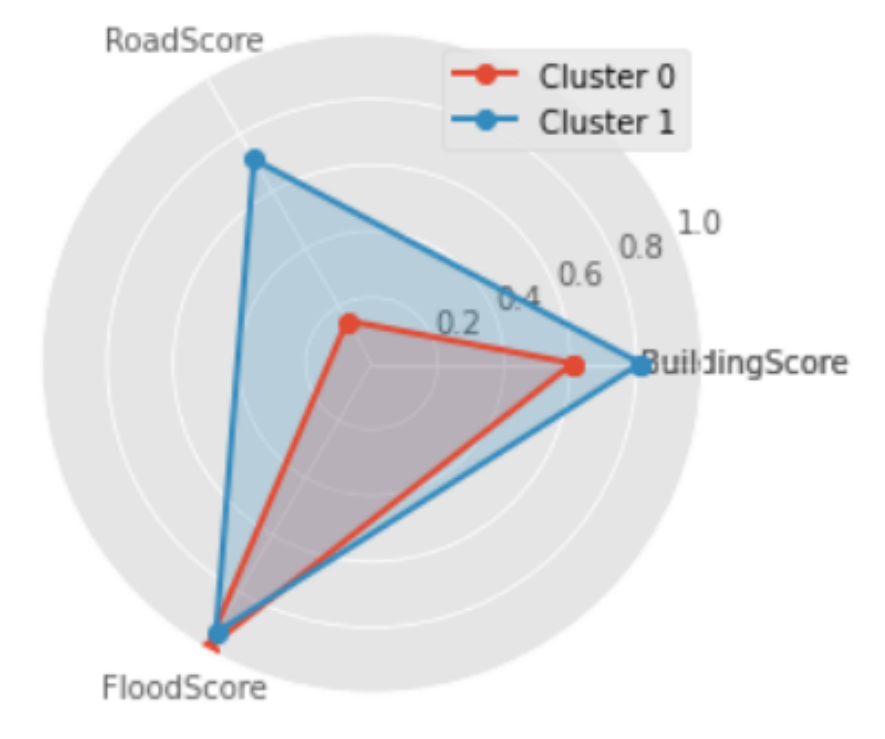}
\captionsetup{justification=centering}
\caption{SOTA model}
\end{subfigure}
\caption{Clustering results of the Baseline model and the SOTA model. (a) Baseline model, (b) SOTA model.}
\label{fig:cluster_baseline_sota}
\end{figure}

\begin{figure}[t]
\centering
\includegraphics[width=\textwidth]{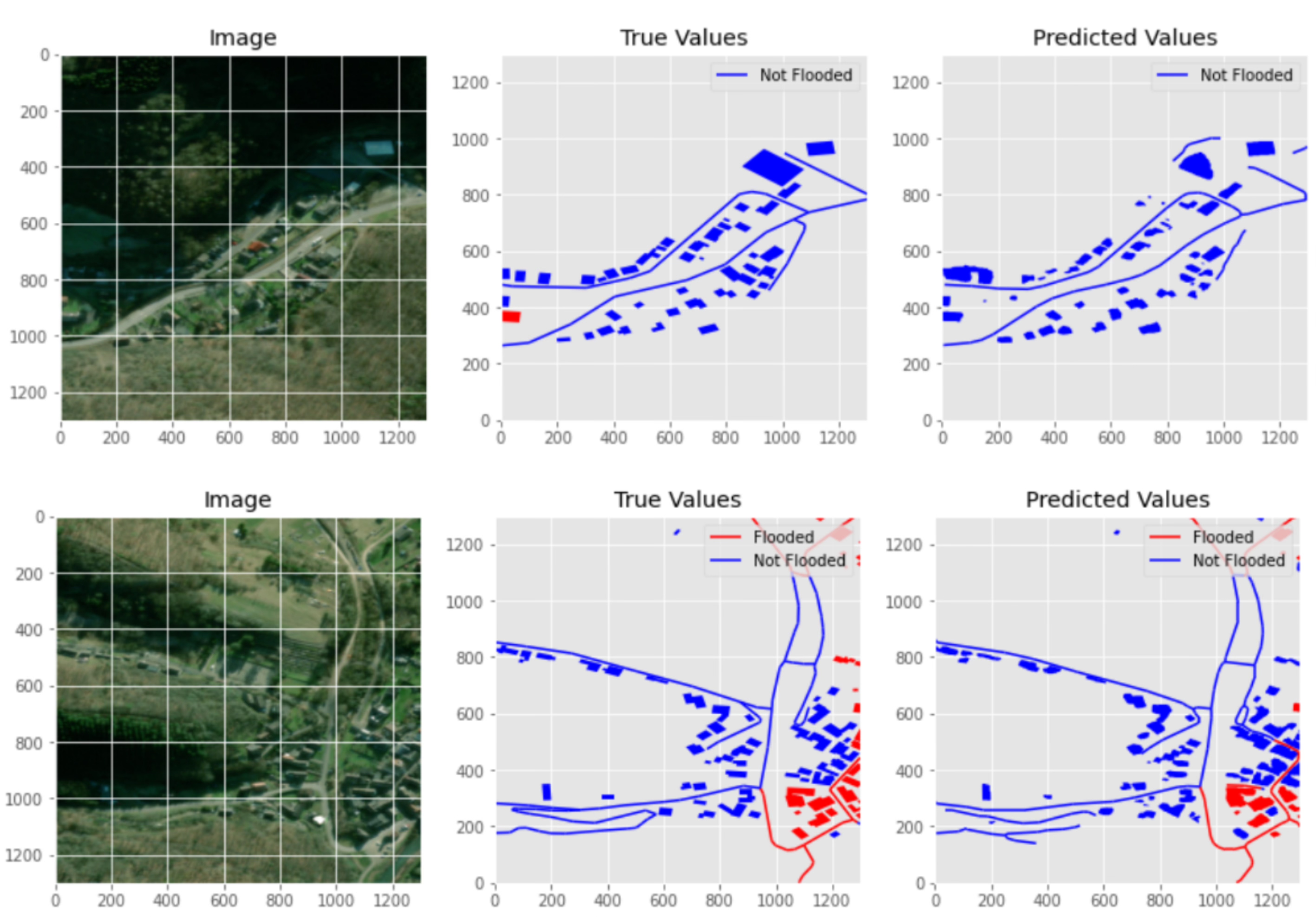}
\caption{Comparison between satellite images, annotated images, and SOTA prediction images. From top to bottom are examples of \textbf{Cluster 0}, \textbf{Cluster 1}.}
\label{compar_cluster_sota}
\end{figure}

Through observation, we identify two distinct clusters, as shown in Figure \ref{fig:cluster_baseline_sota}. In \textbf{Cluster 1}, all metrics exhibit outstanding performance, showcasing excellent features. Conversely, in \textbf{Cluster 0}, while flood prediction is relatively accurate, scores for the other indicators are slightly lower. Nonetheless, as shown in Figure \ref{compar_cluster_sota}, the SOTA model demonstrates overall favorable predictive performance.

\subsection{Experiment 2: Bad Case Analysis}
In our second experiment, we analyze the predictive results of the Baseline and SOTA models to pinpoint cases where performance is lacking. By setting specific criteria for bad cases, we aim to capture a comprehensive view of instances where the models fall short. This analysis is crucial for understanding the limitations of our models and for guiding future improvements.
We analyze the predictive results of both Baseline and SOTA models to identify under-performing cases. Firstly, the criterion of bad case 1 includes cases where scores were below their respective median. Secondly, the criterion of bad case 2 means that any case with a score below 25th percentile. By combining these two standards, we can comprehensively capture cases of poor performance.

\subsubsection{Bad Cases of Baseline Model}
\begin{figure}[t]
\centering
\begin{minipage}[b]{0.45\textwidth}
\includegraphics[width=\textwidth]{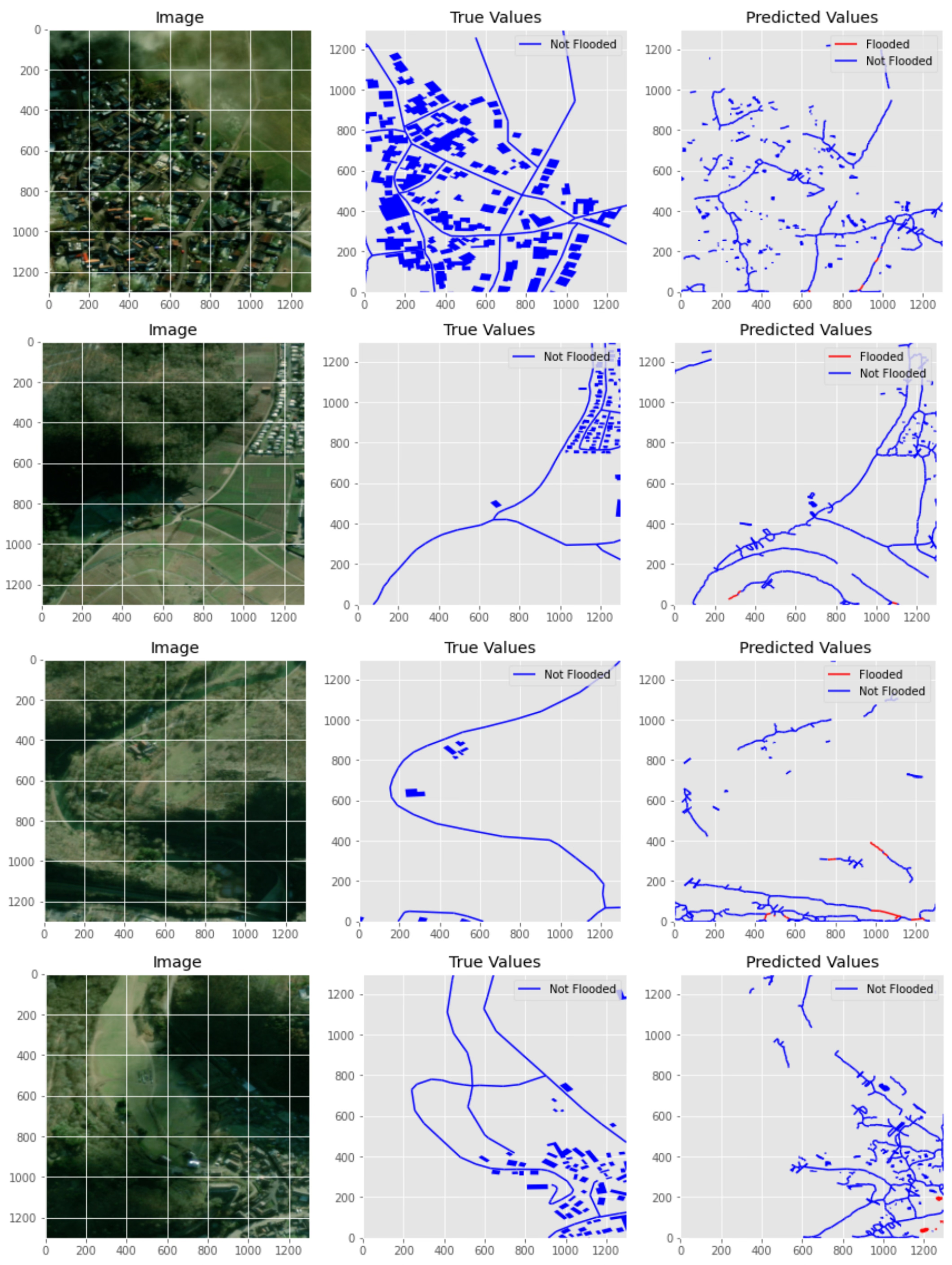}
\end{minipage}
\quad
\begin{minipage}[b]{0.45\textwidth}
\includegraphics[width=\textwidth]{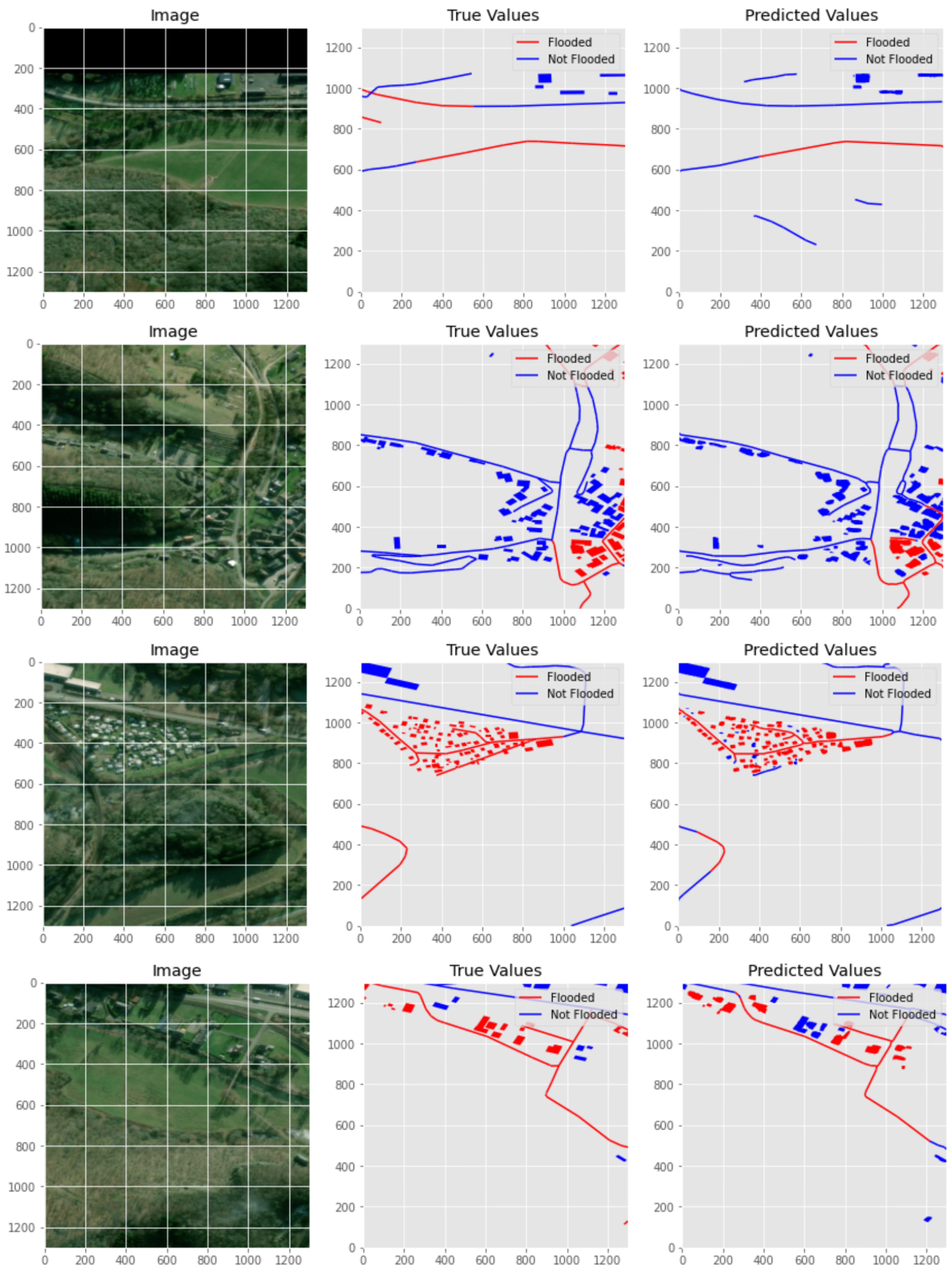}
\end{minipage}
\caption{Left are the bad cases of the Baseline model, from top to bottom are bad case 1, bad case 2, bad case 3, bad case 4. Right are the bad cases of the SOTA model, from top to bottom are bad case 5, bad case 6, bad case 7, bad case 8.}
\label{bad_cases}
\end{figure}

As shown in Figure \ref{bad_cases}, bad case 1 shows a deficiency in the Baseline model's ability to identify buildings. It exhibits issues with the continuity of road identification, with roads appearing disjointed. Additionally, the distribution of buildings is sparse, differing significantly from the block-like buildings in the annotations. This underscores shortcomings in the Baseline method's accuracy in identifying building outlines and the continuity of roads. Bad case 2 shows an error in the Baseline model's judgment of flood occurrences. It mistakenly labels the presence of flooding in the lower-left portion of the image, where no flooding actually occurs (although this may also be attributed to incomplete annotation by the annotators regarding all roads on the image). Additionally, unclear road identification and incorrect labeling of certain buildings are also evident. Bad case 3 shows the primary issue with the Baseline method, i.e., inconsistent road labeling. Despite the main road being clearly visible, the presence of shadows leads to highly fragmented road identification. Moreover, in other shadowed areas, there are errors in road recognition and incorrect flood assessments. It is apparent that under conditions of reduced visibility (such as shadows or fog), the Baseline method's accuracy in road identification and flood assessment is significantly compromised. Bad case 4 shows that, in addition to the previously mentioned issues, the Baseline model struggles with identifying complex building clusters even under highly clear conditions. There is a significant disparity between the model's recognition of complex buildings and the actual structures, resulting in inaccurate differentiation between buildings and determining whether they are affected by floods.

Combining the aforementioned four cases, it is evident that the Baseline model exhibits certain deficiencies in both the continuity of road identification and the delineation of building areas. In the Figure \ref{compar_cluster_sota},we provide a visual comparison between satellite images, annotated images, and the predictions made by the SOTA model for flood detection. The images are presented in a sequence from top to bottom, showcasing examples from two distinct clusters: Cluster 0 and Cluster 1. The satellite images serve as the base, the annotated images indicate the ground truth with marked flood-affected areas, and the SOTA prediction images display the model's interpretation of flood events in the corresponding scenes. This comparison aims to highlight the model's performance in accurately identifying flooded regions across different clusters.

\subsubsection{Bad Cases of SOTA Model}
Bad case 5 in Figure \ref{bad_cases} shows the precision of the SOTA model in detecting flood events, albeit with instances of missed detection. Moreover, while the building annotations are accurate in this case, there are mislabeled road segments that do not correspond to actual roads, indicating some shortcomings in road identification. Bad case 6 has an overall low score, but its annotation of building outlines is remarkably accurate, which is a characteristic of the SOTA method. The labeling of roads and buildings is mostly precise, however, there are errors in identifying some buildings as flood-affected areas, indicating certain deficiencies in flood determination. Bad case 7 exhibits highly accurate annotations of buildings, but it demonstrates some misjudgments in flood recognition within the central area and also misses two flood-affected segments in the lower-left part regarding road flood assessment. Bad case 8 similarly shows errors in determining flood-affected areas, as it overlooks an entire block of flood-affected buildings. 

\begin{table}[t]
\centering
\caption{Quantitative comparison of performance using different improvement methods on baseline and Top1 models.}
\label{tab:comparison}
\small 
\renewcommand{\arraystretch}{1.2} 
\setlength{\tabcolsep}{0.3mm} 
\begin{tabular}{ccccccc}
\hline
\textbf{Model} & \textbf{Improv.} & \textbf{Prec. (\%)} & \textbf{Recall (\%)} & \textbf{F1-Score (\%)} & \textbf{IoU (\%)} & \textbf{Acc. (\%)} \\
\hline
Baseline & --- & 0.758 & 0.812 & 0.784 & 0.645 & --- \\
Baseline & Err. Data Rem. & 0.789 & 0.821 & 0.805 & 0.674 & --- \\
Baseline & Hist. Eq. & 0.799 & 0.777 & 0.788 & 0.65 & --- \\
\hline
Top1 Model & --- & --- & --- & --- & 0.996 & 0.727 \\
Top1 Model & Err. Data Rem. & --- & --- & --- & 0.996 & 0.749 \\
\hline
\end{tabular}
\end{table}

Combining the aforementioned four cases, it can be observed that the SOTA model performs well in identifying roads and building areas, accurately delineating road lines and building regions. However, there are certain issues in identifying flood-affected buildings.

\subsection{Experiment 3: Manual intervention}
The third experiment focuses on manual intervention, particularly in filtering and removing problematic data from our dataset. Through a thorough analysis of errors, we aim to eliminate inaccuracies in labeling that could potentially skew the evaluation metrics of our models. By refining the dataset, we seek to ensure that the true effectiveness of our models is accurately reflected, thereby enhancing their predictive performance.

\subsubsection{Filtering and Removing Problematic Data}
Through a comprehensive analysis of errors in the dataset, we observed that several images contained labeling inaccuracies, including missing labels, incorrect labels, and compromised image integrity. The presence of these inaccuracies inevitably lowers the ceiling of model evaluation metrics, thereby influencing the assessment of the model's performance. To pinpoint these error cases, this research employed the model developed in Experiment 1 to predict outcomes for the entire dataset, encompassing both the training and validation sets. Priority was given to examining samples with lower metrics, followed by those with higher metrics. This process led to the identification of 32 instances of incorrect labeling, primarily concerning the labeling of buildings, as delineated in the subsequent table. These samples exhibited clear labeling issues and were deemed unsuitable for further training or validation purposes. Upon the exclusion of the data with identified errors, this study proceeded to retrain and validate both the Baseline model and the Top1 model. It was discernible that the elimination of erroneous data resulted in a significant enhancement of the performance metrics for each model. This outcome suggests that the true effectiveness is more accurately reflected when the influence of erroneous data is mitigated.

\subsubsection{Preprocessing with Contrast Enhancement via Histogram Equalization}
Histogram equalization is a common technique used to enhance the contrast of an image by modifying the distribution of the image’s histogram and it is found to be very effective especially in rows of the histogram where the contrast is generally low. The method is used to improve the quality of various images in many disciplines including the image data presented in this document. The following steps highlight how histogram equalization works: 

First,the histogram of the original geospatial image is computed. A histogram is a plot that shows the frequency distribution of every grayscale level in an image. For grayscale representation, the histogram shows the value of the frequency of occurrence of each grayscale level.

Second, the computation of the cumulative histogram: The cumulative histogram, denoted as \( C(i) \), is the cumulative sum of the histogram, representing the total count of pixels for all grayscale levels up to and including level \( i \).

\begin{equation}
C(i) = \sum_{j=0}^{i} H(j)
\end{equation}

Third, the process of normalizing the cumulative histogram involves adjusting its values to fall within the interval [0, 1]. This is achieved by dividing each value in the cumulative histogram by the total number of pixels, N, in the image. The normalized cumulative histogram is represented as follows:

\begin{equation}
C^\prime(i) = \frac{C(i)}{N}
\end{equation}

Fourth, the process of mapping to new grayscale levels involves the remapping of the original image's pixel values to new grayscale levels. This is achieved by utilizing the normalized cumulative histogram, which serves as the basis for this transformation. The remapping process follows a specific formula, where 'L' denotes the number of possible grayscale levels. In the case of an 8-bit image, for example, 'L' would be equal to 256. 

\begin{equation}
h(v) = \text{round}\left((L - 1) \cdot C^\prime(v)\right)
\end{equation}

\begin{figure}[t]
\centering
\begin{subfigure}[t]{0.45\linewidth}
\includegraphics[width=\linewidth]{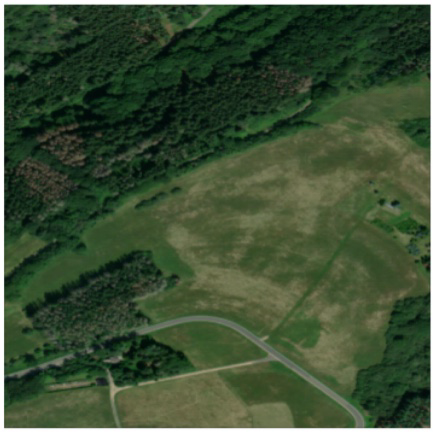}
\caption{True label}
\label{fig:comparison_a}
\end{subfigure}
\quad 
\begin{subfigure}[t]{0.45\linewidth}
\includegraphics[width=\linewidth]{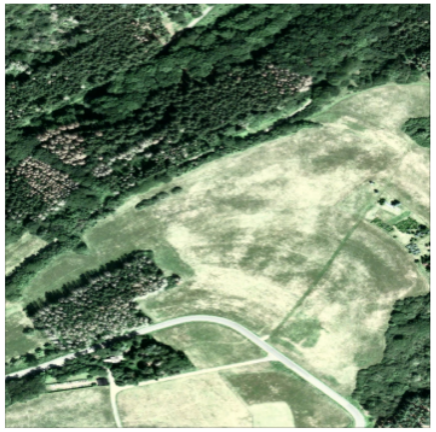}
\caption{Pred label}
\label{fig:comparison_b}
\end{subfigure}
\caption{Comparison Before and After Histogram Equalization}
\end{figure}

Through histogram equalization, gray levels in the original image are redistributed to create a more uniform histogram, enhancing image contrast. This process disperses pixels across the entire gray scale, increasing the dynamic range. Consequently, dark regions become brighter, and details are clarified, while bright areas retain their luminosity. However, excessive enhancement can result in unnatural visual effects, especially in areas with high or low pixel concentration. Therefore, adaptive histogram equalization methods, such as contrast-limited adaptive histogram equalization, are necessary in practical applications to avoid over-enhancement. Based on histogram equalization techniques, this paper enhanced the contrast of image data, with the specific enhancement effect shown in the Figure \ref{fig:comparison_b}. The left image is the original image, and the right image is the image after contrast enhancement. It can be observed that after histogram equalization processing, the brightness of the image is more uniform, and the contrast is significantly enhanced, theoretically making it more suitable for model training.

The Figure \ref{fig:comparison_b} display the Comparison Before and After Histogram Equalization:(a) is the original image before histogram equalization; (b) is the processed image after histogram equalization, where an evident enhancement in image contrast can be observed. Following the application of histogram equalization, this study undertook the task of retraining and validating both the baseline model and the Top1 model, with the specific outcomes being presented in the table below. In comparison to the performance metrics prior to contrast enhancement, a notable improvement in the model's Precision was observed subsequent to contrast enhancement. However, this enhancement was accompanied by a slight decline in other metrics, such as Recall. This phenomenon suggests that augmenting image contrast does not unequivocally lead to an enhancement in the training performance of the model, highlighting the nuanced interplay between image processing techniques and model training dynamics.

\section{Discussion} 
This research focuses on the SpaceNet 8 dataset, which integrates remote sensing machine learning training data for building footprint detection, road network extraction, and flood detection. Our study analyzes and predicts the impact of floods on buildings and roads using Baseline and SOTA models. Apache Sedona analysis reveals misclassifications in specific scenarios, indicating room for improvement in existing models. Common misclassifications highlight the need for better data annotation accuracy. Future research should prioritize data quality and annotation accuracy to enhance model prediction performance.

This study demonstrates the application of error analysis in extracting flood-damaged buildings and roads using Apache Sedona and clustering algorithms for error analysis. By leveraging Apache Sedona's geospatial capabilities and clustering algorithms, this approach effectively predicts flood impacts on infrastructure from historical data. This method enhances problem-solving efficiency and accuracy and provides guidance for future research.

Future research can explore more efficient spatial data algorithms and improve data annotation workflows to enhance prediction performance. Additionally, integrating advanced technologies like deep learning could further improve spatial data analysis and forecasting. Apache Sedona's spatial data processing and analysis hold promising prospects for future research.

\section{Conclusion}
In this study, we utilized datasets from the SpaceNet8 Challenge to focus on the semantic segmentation of satellite imagery for detecting buildings and roads before and after flood events. We explored error analysis on the Apache Sedona platform and iterative model improvement, where Apache Sedona, an open-source geospatial data analysis engine, provided significant support for processing satellite imagery. Through visual analysis, we identified two main error types in the model: annotation errors in the data and errors due to low image contrast. To address these, we carefully examined all training and validation data, removing instances of annotation errors and applying histogram equalization to enhance image contrast. However, the latter did not significantly impact the model's performance, possibly because flood disaster monitoring relies more on geographical location information and the spatial layout of buildings than on contrast.

Our research applied the Apache Sedona platform to extract flood-damaged buildings and roads, employing clustering algorithms for error analysis. This showcased the application of error analysis, a method that solves new problems by referencing similar past cases\cite{darias2022using}, across aspects like case retrieval, adaptation, revision, and retention. This approach enhanced problem-solving efficiency and accuracy, offering methodological references for future research. Our work marginally contributes to satellite image detection of flood disasters, particularly for the SpaceNet8 Challenge, and provides insights into  the application of error analysis the application in disaster monitoring.


\bibliography{references}
\bibliographystyle{splncs04}

\end{document}